\documentclass[sigconf]{aamas}  %

\usepackage{booktabs}
\usepackage{amsmath}
\usepackage{gensymb}
\usepackage{algorithm}
\DeclareMathOperator*{\argmin}{\arg\min}   %

\usepackage{bm}
\usepackage[load-configurations=version-1]{siunitx} %
\usepackage{algorithmic}

\usepackage{xcolor}

\newcommand{\norml}{{NoRML}} %
\newcommand{\maml}{{MAML}}

\newcommand{\advparams}{\bm{\psi}}
\newcommand{\adv}{A_{\advparams}}

\newcommand{\offparams}{\bm{\theta}_\text{\small offset}}
\newcommand{\observation}{\bm{s}}
\newcommand{\action}{\bm{a}}
\newcommand{\transition}{(\observation_t, \action_t, \observation_{t+1})}
\usepackage{balance}

\usepackage{subcaption}
\setcopyright{ifaamas}  %
\acmConference[AAMAS'19]{Proc.\@ of the 18th International Conference on Autonomous Agents and Multiagent Systems (AAMAS 2019)}{May 13--17, 2019}{Montreal, Canada}{N.~Agmon, M.~E.~Taylor, E.~Elkind, M.~Veloso (eds.)}  %
\acmYear{2019}  %
\copyrightyear{2019}  %

\begin{document}

\title{\norml{}: No-Reward Meta Learning}  %

\author{Yuxiang Yang, Ken Caluwaerts, Atil Iscen, Jie Tan, Chelsea Finn}
\affiliation{%
  \institution{Robotics at Google}
  \postcode{10019}
}
\email{{yxyang, kencaluwaerts, atil, jietan, chelseaf}@google.com}

\begin{abstract}  %
Efficiently adapting to new environments and changes in dynamics is critical for agents to successfully operate in the real world. Reinforcement learning (RL) based approaches typically rely on external reward feedback for adaptation. However, in many scenarios this reward signal might not be readily available for the target task, or the difference between the environments can be implicit and only observable from the dynamics. To this end, we introduce a method that allows for self-adaptation of learned policies: No-Reward Meta Learning (\norml{}). \norml{} extends Model Agnostic Meta Learning (\maml{}) for RL and uses observable dynamics of the environment instead of an explicit reward function in \maml{}'s finetune step. Our method has a more expressive update step than \maml{}, while maintaining \maml{}'s gradient based foundation. Additionally, in order to allow more targeted exploration, we implement an extension to \maml{} that effectively disconnects the meta-policy parameters from the fine-tuned policies' parameters. We first study our method on a number of synthetic control problems and then validate our method on common benchmark environments, showing that \norml{} outperforms \maml{} when the dynamics change between tasks. 

\noindent Videos and source-code are available at \url{https://sites.google.com/view/noreward-meta-rl/}.
\end{abstract}


\maketitle

\section{Introduction}

Adapting to new environments is a crucial capability for autonomous robots to operate in the real world. For example, after a robot learns to walk, its dynamics may change due to hardware failures, its sensor measurements (e.g. IMU) may drift over time, and more importantly, the reward signals may no longer be available due to lack of corresponding sensors after the robot is deployed (e.g. a tracking system that measures walking distance). How can the robot learn to adapt even after these changes? 

Model Agnostic Meta Learning (\maml{}) \cite{DBLP:journals/corr/FinnAL17} tackles the above problem by training a \emph{meta-policy} that is optimized for quick adaptation to new tasks. During adaptation, this meta-policy can be fine-tuned efficiently with a small amount of data that is collected in the new environment. While \maml{} is successful at adapting policies to different tasks that are defined by reward changes (e.g. running forward vs. backward), it is less effective when adapting to other changes \cite{finn2018thesis}, such as dynamics changes, sensor drifts, or missing reward signals. 

In this paper, we introduce No-Reward Meta Learning (\norml{}) to address the above challenges. The key insight underlying \norml{} is that we can simultaneously learn the meta-policy and the advantage function used for adapting the meta-policy, optimizing for the ability to effectively adapt to varying dynamics. The meta-learned advantage function serves two purposes. First, it allows the policy to be adapted without external rewards: even if the reward signal is not present during the adaptation stage, we can use the learned advantage function to evaluate the current policy and compute the gradient. Second, the learned advantage gives more expressive power to the meta-optimization over how the policy can be updated, since it can learn to implicitly shape its reward feedback. As a result, the meta-policy can better recognize and adapt to subtle dynamics changes.

Beyond learning to update the policy parameters, one critical aspect of the meta-RL problem is the sampling distribution induced by the meta-policy, i.e. how the meta-policy chooses to explore and collect trajectory samples that are maximally informative about the unknown task or environment~\cite{stadie2018some,gupta2018meta}. In the original \maml{} formulation, the same policy parameters are used for collecting experience as for the gradient computation, limiting the extent to which the meta-policy parameters can be used for each individual purpose. To further increase the expressive power of the meta-optimization, we propose to decouple these two roles by augmenting the adaptation process with a meta-learned parameter offset. The parameter offset is added to the initial \maml{} parameters after sampling and during gradient adaptation, allowing different parameter vectors to be used for each while defaulting to the case where they are the same, akin to how residual networks (ResNets)~\citep{he2016identity} default to the identity function.
With the combination of the offset policy and the learned advantage function, \norml{} can successfully adapt to changes in dynamics, sensor drifts, and missing reward signals.

We evaluate \norml{} on three control domains with varying sources of dynamics changes: an illustrative point agent example with disoriented actions, a cartpole with sensor bias, and a half-cheetah with wiring errors. In comparison to MAML, \norml{} enables adaptation from a single trial, does not require reward signal for adaptation, and in most cases even leads to improved asymptotic performance.
We find that both the learned advantage function and the parameter offset are important for good performance.

\section{Related Work}
Algorithms for learning to learn~\citep{schmidhuber1987evolutionary,bengio1992optimization,thrun1998learning,hochreiter2001learning}, or meta learning, aim to acquire a procedure that can more efficiently and effectively learn to solve new tasks. We consider meta learning in the context of reinforcement learning, i.e. meta reinforcement learning~\cite{wang2016learning,duan2016rl}. Prior model-free meta reinforcement learning algorithms can generally be categorized as being recurrence-based~\citep{wang2016learning,duan2016rl,mishra2018simple,stadie2018some}, gradient-based~\citep{DBLP:journals/corr/FinnAL17,gupta2018meta,stadie2018some}, or a hybrid of the two~\cite{sung2017learning,houthooft2018evolved}. We build a gradient-based meta-RL method that extends the MAML algorithm~\citep{DBLP:journals/corr/FinnAL17}. 
Unlike these prior model-free meta-RL works, we focus on the problem of learning to adapt to different dynamics, rather than adapting to different rewards.

Prior model-based RL approaches have considered the problem of learning to adapt to different dynamics through meta-RL~\citep{saemundsson2018meta,clavera2018learning} or through learned priors~\cite{fu2016one}. Using a learned model is suitable when sample efficiency is a concern, but achieves lower asymptotic performance than model-free meta-RL~\citep{clavera2018learning}. Recent work by~\citet{pmlr-v87-clavera18a} used MAML to adapt to different learned dynamics models within an ensemble, to improve model-based RL. We also consider the problem of adapting to different dynamics, but in the context of adapting to different environments, rather than different estimated models of the same environment. Further, our method improves upon MAML by not requiring a reward function for adaptation.

Separate from the meta learning literature, many other works have considered the problem of adapting to varying dynamics through, e.g. adaptive inverse control~\citep{widrow1990adaptive}, self-modeling~\citep{bongard2006resilient}, Bayesian optimization~\citep{cully2015robots}, or online system identification~\citep{ljung1998system,yu2017preparing}. Our approach leverages prior experience to learn to adapt a policy with as little as a single trial without observed reward, and has few assumptions about the nature of the dynamics changes.
Other methods have leveraged prior experience to learn a single policy that is robust to many different dynamics~\citep{mordatch2015ensemble,rajeswaran2016epopt,tan2018sim,peng2018sim}. Our experiments illustrate several realistic scenarios where robustness is not sufficient and adaptation is critical to good performance.

\section{Preliminaries}
In this section, we overview model-free reinforcement learning, describe gradient-based meta learning (\maml{}), and discuss the potential difficulties that the vanilla \maml{} algorithm can have in model-free RL scenarios. We also introduce notation. %

\subsection{Model-free Reinforcement Learning}
We study reinforcement learning problems where the agent makes a sequence of actions in a stochastic environment in order to maximize the cumulative reward. Formally, we define the problem as a Markov decision process (MDP) which consists of: a state space $\mathcal{S}$, an action space $\mathcal{A}$, a transition probability distribution $p(\observation_{t+1}|\observation_t, \action_t)$, a reward function $R:\mathcal{S}\times\mathcal{A}\rightarrow\mathbb{R}$ and an initial state distribution $p(\observation_1)$.%

In model-free RL, we aim to directly optimize a policy $\pi_\theta:\mathcal{S}\rightarrow \mathcal{P}(\mathcal{A})$, where $\mathcal{P}(\mathcal{A})$ is the set of probability distributions on the action space, and $\bm{\theta}\in\mathbb{R}^n$ is a vector to parameterize the policy. The agent interacts with the environment over a finite horizon of length $H$ and collects a trajectory $D = \{\observation_1, \action_1, r_1, \ldots, \observation_{H+1}\}$ over $\mathcal{S}\times\mathcal{A}\times\mathcal{R}$. The agent tries to find policy parameters $\bm{\theta}$ that maximize the expected return. Equivalently, we can instead minimize the expected loss, which can be written as:
\begin{equation}
    \begin{aligned}
        \mathcal{L}(\bm{\theta}, D)=-\mathbb{E}_{(\bm{s}_t,\bm{a}_t)\sim\pi_{\bm{\theta}}}
        \left[\sum_{t=1}^H R(\bm{s}_t,\bm{a}_t)\right] = 
        -\mathbb{E}_{(\bm{s}_t,\bm{a}_t)\sim\pi_{\bm{\theta}}}\left[\sum_{t=1}^H r_t\right].  
    \end{aligned}
\end{equation}

\textit{Policy gradients} is a popular model-free algorithm to optimize a policy. The algorithm approximates the gradient  $\nabla_{\bm{\theta}}\mathcal{L}(\bm{\theta}, D)$ using the \textit{Policy Gradient Theorem}~\cite{sutton2000policy}:
\begin{equation}\label{policy_gradient}
    \begin{aligned}
        \nabla_\theta \mathcal{L}(\bm{\theta}, D)=-\mathbb{E}_{(\bm{s}_t, \bm{a}_t)\sim\pi_\theta}\left[A^{\pi}(\bm{s}_t, \bm{a}_t)\nabla_{\bm{\theta}}\log\pi_{\bm{\theta}}(\bm{a}_t|\bm{s}_t)\right],
    \end{aligned}
\end{equation}
The expectation in Eq.~\ref{policy_gradient} is computed by Monte-Carlo estimation. where we replaced the reward signal $r_t$ with the advantage function $A^{\pi}(\bm{s}_t, \bm{a}_t)$. 
To reduce the variance of the gradient estimations, an advantage function of the following form can be used:
\begin{equation}
    A^{\pi}(\bm{s}_t,\bm{a}_t)=\sum_{t'=t}^H\gamma^{t'-t} r_{t'}-V^{\pi}(\bm{s}_t),
\end{equation}
where $V^{\pi}(\bm{s})$ is a fitted value function estimator (also called \textit{critic}) and $\gamma$ is a discount factor.

\subsection{Gradient-based Meta Learning}
Meta learning algorithms optimize for a learning procedure that can quickly adapt to a particular task. Assuming that the training and testing tasks share some commonalities and are sampled from the same distribution, meta learning algorithms aim to learn the structure underlying the tasks and use this knowledge for fast learning. The meta-training process usually involves drawing data from different tasks and optimizing for performance after learning with a small amount of data.

Model-agnostic meta learning (\maml{})~\cite{DBLP:journals/corr/FinnAL17} takes a gradient-based approach to the above problem.
Formally, given a distribution over tasks $p(\mathcal{T})$, where each task defines a specific loss function $\mathcal{L}_{\mathcal{T}_i}$, \maml{} aims to find a good meta parameters %
$\theta$ that, with one step of gradient descent, can adapt to specific tasks with small amounts of data. The objective can be described as follows:
\begin{equation}\label{MAML_loss}
    \begin{aligned}
        \argmin_{\bm{\theta}}\mathbb{E}_{\mathcal{T}_{i}\sim p(\mathcal{T})}\left[\mathcal{L}_{\mathcal{T}_{i}}(\bm{\theta}_{i})\right] \text{ s.t. }\bm{\theta}_{{i}}=\bm{\theta} - \alpha\nabla_{\bm{\theta}}\mathcal{L}_{\mathcal{T}_{i}}(\bm{\theta}).
    \end{aligned}
\end{equation} 

To find a good set of meta parameters,
 $\bm{\theta}$, \maml{} uses gradient descent on the meta objective in Eq.~\ref{MAML_loss}. This requires second-order derivatives w.r.t. $\bm{\theta}$.
When presented with data for a new test task $\mathcal{T}_j$, \maml{} adapts by simply performing one step of gradient descent starting from $\bm{\theta}$. 
Since the formulation of \maml{} is quite general, it can be applied to a range of problems, including model-free RL, which we describe next.

\subsection{\maml{} on Model-free RL}

Applying \maml{} in the context of model-free reinforcement learning (\maml{}-RL), %
$\bm{\theta}$ parameterizes the \textit{meta-policy} $\pi_{\bm{\theta}}$. To perform task-specific fine-tuning for a test task, one collects trajectories (\textit{meta rollouts}) using the meta-policy $\pi_{\bm{\theta}}$ on a sampled task $\mathcal{T}_i$ and then uses the policy gradient equations (Eq.~\ref{policy_gradient}) to obtain the \textit{fine-tuned} (also called \textit{adapted}) policy $\pi_{\bm{\theta}_i}$. 
We can write the adaptation step for task $\mathcal{T}_i$ explicitly by applying the gradient update rule in Eq.~\ref{MAML_loss} to model-free RL and replacing the expectation with Monte-Carlo estimation. More precisely, we collect $K$ trajectories $D_i^{\small\mbox{train}}=\{(\bm{s}_1, \bm{a}_1, r_{1}, \bm{s}_2, \bm{a}_2, r_{2} \ldots \bm{s}_{H+1})_k: k=1\ldots K\}$ using the meta-policy $\pi_{\bm{\theta}}$ on task $\mathcal{T}_i$ and approximate the policy gradient:
\begin{align}\label{maml_pg}
    \bm{\theta}_i& =\bm{\theta}-\alpha\nabla_{\bm{\theta}}\mathcal{L}_{\mathcal{T}_i}(\bm{\theta}, D_i^{\small\mbox{train}})\\
                & = \bm{\theta}+\alpha\!\!\!\!\!\!\!\!\sum_{(\observation_t,\action_t,r_t) \in D_i^{\mbox{\tiny train}}}\!\!\!\!\!\!\!\! A^{\pi}(\observation_t, \action_t)\nabla_{\theta}\log\pi_\theta(\action_t|\observation_t). \label{maml_pg_explicit}
\end{align}

During meta training, one also collects rollouts of the fine-tuned policies $\bm{\theta}_i$ for $\mathcal{T}_i$ to compute the meta objective in Eq.~\ref{MAML_loss}.
As stated in Eq.~\ref{MAML_loss}, \maml{} optimizes the meta-policy parameters $\bm{\theta}$ such that the expected loss over all tasks \textit{after} adaptation is minimized. To this end, we use $K$ trajectories $D_i^{\small\mbox{test}}$ from the adapted policies $\pi_{\theta_i}$ to approximate the gradient of the meta-objective: 
\begin{equation}\label{maml_meta_pg}
    \bm{\theta} \leftarrow \bm{\theta} -\beta\sum_{\mathcal{T}_i\sim\mathcal{T}}\nabla_{\bm{\theta}}\mathcal{L}_{\mathcal{T}_i}(\bm{\theta}_i,  D^{\small\mbox{test}}_i),
\end{equation}
where 
\begin{equation}
    \mathcal{L}_{\mathcal{T}_i}(\bm{\theta}_i, D^{\small\mbox{test}}_i) = {\mathcal{L}_{\mathcal{T}_i}(\bm{\theta}-\alpha\nabla_{\bm{\theta}}\mathcal{L}_{\mathcal{T}_i}(\bm{\theta}, D_i^{\small\mbox{train}}), D^{\small\mbox{test}}_i)}.
\end{equation}
The gradient of $\mathcal{L}_{\mathcal{T}_i}(\bm{\theta}_i, D^{\small\mbox{test}}_i)$ can be approximated using the policy gradient algorithm (c.f. Eq.~\ref{maml_pg_explicit}).

The complete \maml{} algorithm for reinforcement learning is listed in Algorithm~\ref{algo:maml}. 
Algorithm~\ref{algo:maml-adaptation} lists how to perform fine-tuning on a specific task, based on an optimized meta-policy.
Unlike the original MAML-RL study \cite{DBLP:journals/corr/FinnAL17} where different tasks corresponded to different rewards, our goal is to handle the setting where a different task entails a change of the environment, including both dynamics changes and sensor drifts, and where rewards are not present during adaptation, a scenario that MAML-RL cannot handle effectively.

\begin{algorithm}
\caption{\maml{} Training\label{algo:maml}~\cite{DBLP:journals/corr/FinnAL17}}
\begin{algorithmic}
\REQUIRE{$p(\mathcal{T})$: task distribution}
\REQUIRE{$\alpha$: adaptation learning rate} 
\REQUIRE{$\beta$: meta learning rate}
\STATE{Randomly initialize $\bm{\theta}$}%
\WHILE{not done} 
\STATE{
    Sample a batch of tasks $\mathcal{T}_i \sim p(\mathcal{T})$
    \FORALL {$\mathcal{T}_i$ in batch} 
    \STATE{
        Sample $K$ trajectories $D_i^{\tiny\mbox{train}}$ using $\pi_{\bm{\theta}}$ on task $\mathcal{T}_i$. \\
        Compute adapted parameters using $D_i^{\tiny\mbox{train}}$:\\
        $\bm{\theta}_i = \bm{\theta}-\alpha\nabla_{\bm{\theta}}\mathcal{L}_{\mathcal{T}_i}(\bm{\theta}, D_i^{\tiny\mbox{train}})$.\\
        Sample $K$ trajectories $D_i^{\tiny\mbox{test}}$ using $\pi_{\bm{\theta}_i}$ on task $\mathcal{T}_i$. \\
    } 
    \ENDFOR \\
    Update $\bm{\theta} $%
    \,using all $D^{\tiny\mbox{train}}_i$, and $D^{\tiny\mbox{test}}_i$: \\
    $\bm{\theta} \leftarrow \bm{\theta}  -\beta\sum_{\mathcal{T}_i\sim\mathcal{T}}\nabla_{\tiny\bm{\theta}}\mathcal{L}_{\mathcal{T}_i}(\bm{\theta}_i,  D^{\tiny\mbox{test}}_i)$\\
    with $\mathcal{L}_{\mathcal{T}_i}(\bm{\theta}_i, D^{\tiny\mbox{test}}_i) = {\mathcal{L}_{\mathcal{T}_i}(\bm{\theta}-\alpha\nabla_{\bm{\theta}}\mathcal{L}_{\mathcal{T}_i}(\bm{\theta}, D_i^{\tiny\mbox{train}}), D^{\tiny\mbox{test}}_i)}$.
    
} 
\ENDWHILE
\end{algorithmic}
\end{algorithm}

\begin{algorithm}
\caption{\maml{} Fine-tuning\label{algo:maml-adaptation}~\cite{DBLP:journals/corr/FinnAL17}}
\begin{algorithmic}
\REQUIRE{$\mathcal{T}_i$: a new test task}
\REQUIRE{$\bm{\theta}, \alpha$: from \maml{} training}
    \STATE{
        Sample $K$ trajectories $D$ using $\pi_{\bm{\theta}}$ on task $\mathcal{T}_i$. \\
        Compute fine-tuned policy: \\
        $\bm{\theta} \leftarrow \bm{\theta}-\alpha\nabla_{\bm{\theta}}\mathcal{L}_{\mathcal{T}_i}(\bm{\theta}, D)$.\\
    } 

\end{algorithmic}
\end{algorithm}

\section{No-reward Meta Learning}
Consider an agent with a single task (i.e. a fixed reward function) such as running forward. Intuitively, if the dynamics change such as calibration errors or motor malfunctions, an agent should not need reward supervision in order to adapt its behavior: the dynamics change is recognizable from the state-action transitions alone. Similarly, a human can adapt to a new terrain without external reward feedback. However, model-free RL requires such rewards. Our goal is to develop a model-free meta-RL algorithm that can learn to quickly adapt a policy to dynamics changes and sensor drifts \emph{without external rewards}. To do so, the meta learning algorithm needs to develop its own internal notion of reward, to learn to explore in a way that is maximally informative of the current conditions, and to be able to learn to recognize changes in dynamics and adapt appropriately.

In this section, we introduce our meta reinforcement learning algorithm, termed No-Reward Meta Learning (NoRML), that aims to address these challenges. \norml{} consists of two additional components to the original MAML-RL formulation: a learned advantage function that internalizes the reward in a way that allows for reward-free adaptation (which we discuss next), and a learned parameter offset that enables better exploration (which we discuss in Section~\ref{section:offset}). The entire meta-training algorithm and meta-test procedure of \norml{} are summarized in Algorithms \ref{algo:norml} and \ref{algo:norml-adaptation}. Note that we use the term ``change of the environment'' and ``different tasks'' interchangeably in the following discussion.

\subsection{Learned Advantage Function}

We introduce a \textit{learned advantage function} $\adv\transition$ to replace the estimated advantage $A^{\pi}(\observation_t, \action_t)$ in Eq.~\ref{maml_pg_explicit}. The reason is twofold. First, the learned advantage function can be used to evaluate a trajectory even if the reward signal is not present during adaptation. Thus, it solves the problem of missing reward signals. Second, it is a generalized function form for the advantage function, which considerably increases the expressiveness of the policy gradient for fine-tuning, giving the meta-optimization more control over how it can update the policy. $\adv$ is a feed-forward neural network that takes in consecutive states and action $\transition$. We initialize the weights $\bm{\psi}$ of the advantage network randomly and train it end-to-end. More specifically, during the meta-training process, we adjust its weights according to the gradient of the meta objective (See Alg. \ref{algo:norml} for more details). Note that the learned advantage $\adv$ is only used during fine-tuning, while the reward-based advantage $A^{\pi}$ is still used to compute the outer gradient during meta training.

Since $\adv$ takes in $\transition$ as input, this allows $\adv$ to detect changes in the dynamics $p(\observation_{t+1}|\observation_t, \action_t)$, and provide a more informed "evaluation" of the actions, compared to using only $(\observation_t, \action_t)$. An additional benefit of using $\adv$ is that we eliminate the need to estimate the value function to calculate the observed advantage. 
In \maml{}, it is difficult to estimate a value function from only a few roll-outs, limiting the effectiveness of the resulting policy gradient.
$\adv$ directly transforms the policy gradient and can provide accurate information to the fine-tune step even when the sample size is small.

Given the learned advantage function $\adv$, The \maml{} adaptation step for the task $\mathcal{T}_i$ is modified to be the following:
\begin{align}
\bm{\theta}_i&=\bm{\theta}-\alpha\nabla_{\bm{\theta}}\mathcal{L}^{\tiny\mbox{NoRML}}_{\mathcal{T}_i}(\bm{\theta}, D_i^{\small\mbox{train}})\\
 &=  \bm{\theta}+{\alpha\sum_{D_i^{\tiny\mbox{train}}}\adv \transition\nabla_{\bm\theta}\log\pi_{\bm{\theta}}(\bm{a}_t | \bm{s}_t)},
\end{align}
where $D^{\tiny\mbox{train}}_i=\{\transition_k: k=1\ldots K, t=1\ldots H\}$ is generated based on $K$ trajectories using the meta-policy $\pi_{\bm{\theta}}$ on task $\mathcal{T}_i$ (i.e. $D^{\tiny\mbox{train}}_i$ contains $KH$ state transitions).

We would like to emphasize that this learned advantage function differs fundamentally from approximating $A^{\pi}(\observation_t, \action_t)$ using a fitted value function. In the latter case, the learned advantage function is trained to predict the actual observed advantage values $A^{\pi}(\observation_t, \action_t)$ in Eq.~\ref{maml_pg_explicit}. In contrast, our learned advantage function $\adv$ is optimized to \textit{transform} or \textit{reshape} the policy gradient $\nabla_{\bm\theta}\log\pi_{\bm{\theta}}(\bm{a}_t | \bm{s}_t)$ in a way that achieves more effective adaptation in a single fine-tune step. 
As a result, the output of our advantage network can be significantly different from the estimated advantage values (in a sense, $\adv$ is not a true advantage function).

\begin{algorithm}
\caption{\norml{} Training}
\begin{algorithmic}
\STATE{Differences between \maml{} and \norml{} are {\color{blue}highlighted}.}
\REQUIRE{$p(\mathcal{T})$: distribution over tasks}
\REQUIRE{$\alpha,\ \beta$: step size hyperparameters}
\REQUIRE{$\bm{\theta}_{\sigma}$: initial log standard deviation of meta-policy}
\STATE{Randomly initialize $\bm{\theta}_{\mu}$ and set $\bm{\theta}=\left[\bm{\theta}_{\mu}^T\,\, \bm{\theta}_{\sigma}^T\right]^T$}
\STATE{\color{blue}  Randomly initialize $\advparams$} 
\STATE{\color{blue} Initialize $\bm{\theta}_{\tiny\mbox{offset}}$ to $[0 \ldots 0]^T$}
\WHILE{not done} 
\STATE{
    Sample a batch of tasks $\mathcal{T}_i \sim p(\mathcal{T})$
    \FORALL {$\mathcal{T}_i$} 
    \STATE{
        Sample $K$ trajectories {\color{blue}without rewards} using $\pi_{\bm{\theta}}$ on task $\mathcal{T}_i$ and store all state transitions as a {\color{blue}set $D^{\tiny\mbox{train}}_i=\{(\bm{s}_t, \bm{a}_t, \bm{s}_{t+1}): \forall{k < K}, \forall{t\le H}\}$}. \\
        Compute adapted parameters using $D^{\tiny\mbox{train}}_i$:\\
        $\bm{\theta}_i = \bm{\theta}+{\color{blue}\bm{\theta}_{\tiny\mbox{offset}}}-{\color{blue}\alpha\sum_{ D_i^{\tiny\mbox{train}}}\adv \transition\nabla_{\bm\theta}\log\pi_{\bm{\theta}}(\bm{a}_t | \bm{s}_t)}$.\\
        Sample $K$ trajectories $D_i^{\tiny\mbox{test}}$ using $\pi_{\bm{\theta}_i}$ on task $\mathcal{T}_i$. \\
    } 
    \ENDFOR \\
    Update $\bm{\theta}$, $\bm{\theta}_{\tiny\mbox{offset}}$, and $\advparams$ using all $D^{\tiny\mbox{train}}_i$, and $D^{\tiny\mbox{test}}_i$: \\
    $\left[\begin{matrix} \bm{\theta} \\ {\color{blue}\bm{\theta}_{\tiny\mbox{offset}}} \\ {\color{blue}\advparams}  \end{matrix}\right] \leftarrow\left[\begin{matrix} \bm{\theta} \\ {\color{blue}\bm{\theta}_{\tiny\mbox{offset}}} \\ {\color{blue}\advparams}  \end{matrix}\right]-\beta\sum_{\mathcal{T}_i\sim\mathcal{T}}\nabla_{\tiny\left[\bm{\theta}^T  {\color{blue}\bm{\theta}_{\tiny\mbox{offset}}^T} {\color{blue}\advparams}^T  \right]^T}\mathcal{L}_{\mathcal{T}_i}(\bm{\theta}_i,  D^{\tiny\mbox{test}}_i)$\\
    with $\nabla\mathcal{L}_{\mathcal{T}_i}(\bm{\theta}_i,  D^{\tiny\mbox{test}}_i) = \sum_{D_i^{\tiny\mbox{test}}} A^{\pi}(\observation_t, \action_t)\nabla\log\pi_{\bm{\theta}_i}(\action_t|\observation_t)$.
}
\ENDWHILE
\end{algorithmic}
\label{algo:norml}
\end{algorithm}

\begin{algorithm}
\caption{\norml{} Fine-tuning}
\begin{algorithmic}
\REQUIRE{$\mathcal{T}_i$: a task}
\REQUIRE{$\bm{\theta}$, $\bm{\theta}_{\tiny\mbox{offset}}$, $\advparams, \alpha$: from \norml{} training}
    \STATE{
        Sample $K$ trajectories without rewards using $\pi_{\bm{\theta}}$ and store all state transitions as a set $D=\{(\bm{s}_t, \bm{a}_t, \bm{s}_{t+1}): \forall{k < K}, \forall{t<H}\}$. \\
        Compute fine-tuned policy: \\
        $\bm{\theta} \leftarrow  \bm{\theta}+\bm{\theta}_{\tiny\mbox{offset}}-\alpha\sum_{D}\adv(\bm{s}_t, \bm{a}_t, \bm{s}_{t+1})\nabla_{\bm\theta}\log\pi_{\bm{\theta}}(\bm{a}_t|\bm{s}_t)$.\\
    } 
\end{algorithmic}
\label{algo:norml-adaptation}
\end{algorithm}

\subsection{Offset Learning\label{section:offset}}

\begin{figure}[htp]
            \includegraphics[trim={6.4cm 0 5cm 0},clip, width=0.5\textwidth]{{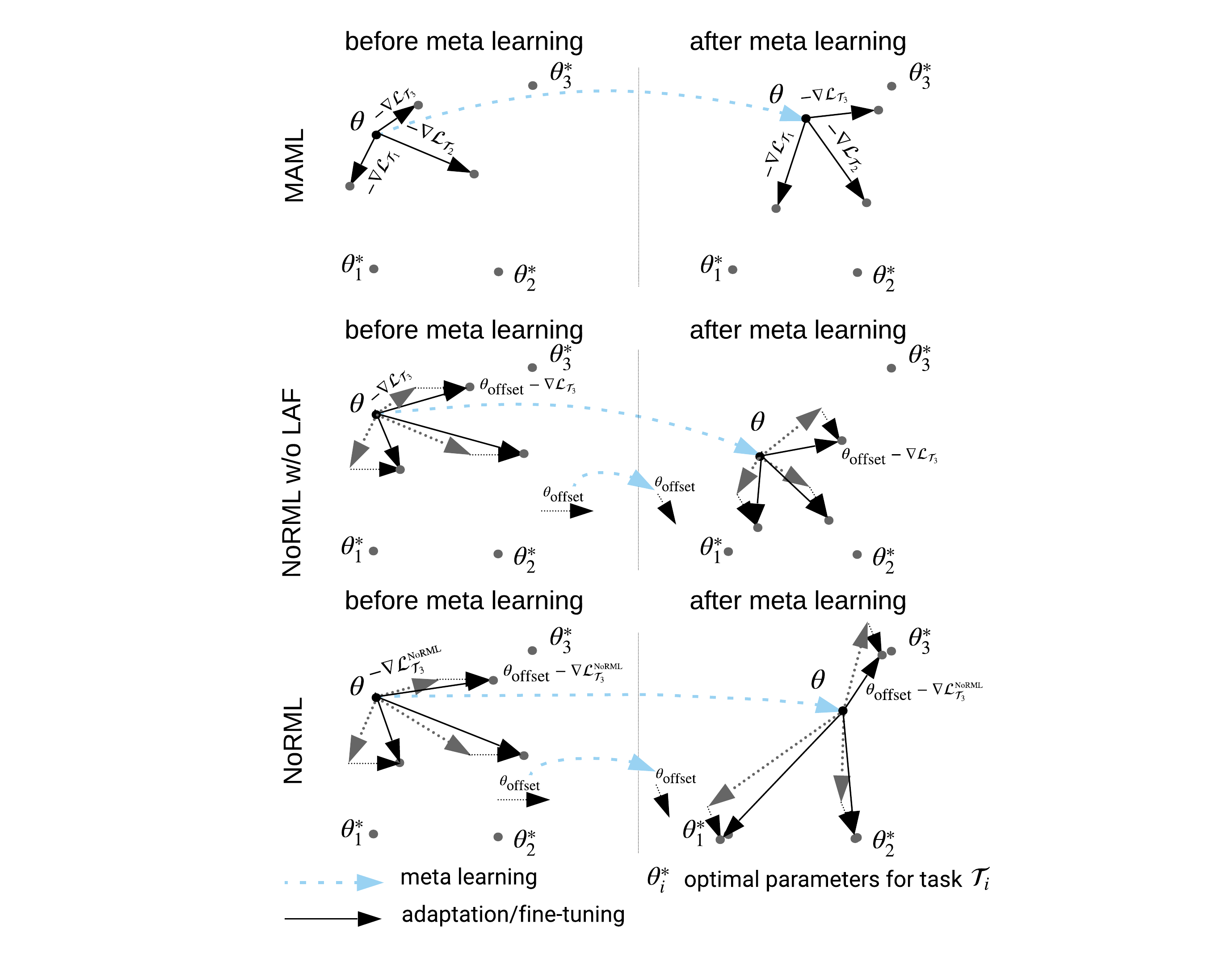}}
            \caption{
            Geometric interpretations of \maml{}, \norml{} w/o Learned Advantage Function (LAF), and \norml{}. \maml{} (top) optimizes a meta-policy $\bm{\theta}$ such that a single fine-tune step using the policy gradient $\nabla_{\bm{\theta}}\mathcal{L}_{\mathcal{T}_i}$ is likely to significantly improve the performance on a specific task $\mathcal{T}_i$. \norml{} without LAF (middle) learns an additional parameter vector $\offparams$. This vector is added to the meta-policy parameters $\bm{\theta}$ and decouples the meta-policy from the fine-tuned policies. 
            \norml{} (bottom) learns an advantage function during meta learning, which results in a modified, more expressive policy gradient for fine-tuning.  
            We also use a policy offset vector $\bm{\theta}_{\small\mbox{offset}}$ with \norml{} to maximize performance. }    
            \label{fig:maml_and_offset}
\end{figure}

Since one policy gradient step may be insufficient to adapt an exploratory meta-policy into a policy for the new task, we introduce a simple, yet effective, technique to decouple the meta-policy from the adapted policies: a learned offset $\bm{\theta}_{\small\mbox{offset}}$ that is added to the policy parameters $\bm{\theta}$ when calculating an adapted policy:
\begin{equation}
    \label{maml_rl_offset}
    \bm{\theta}_i = \bm{\theta}+\bm{\theta}_{\small\mbox{offset}}-{\alpha\sum_{D_i^{\tiny\mbox{train}}}\adv \transition\nabla_{\bm\theta}\log\pi_{\bm{\theta}}(\bm{a}_t | \bm{s}_t)}.
\end{equation}
Note that the policy offset $\bm{\theta}_{\tiny\mbox{offset}}$ is shared for \textit{all tasks}. Hence, it does not influence task-specific adaptation. The adaptation step is still based on trajectories sampled from the meta-policy $\pi_{\bm{\theta}}$, and the policy gradient is still computed with respect to the meta policy parameters $\bm{\theta}$. Similar to the learned advantage function, the parameter offset is optimized end-to-end together with the meta-policy, as shown in Algorithm~\ref{algo:norml}. 
Fig.~\ref{fig:maml_and_offset} shows a geometric interpretation of of \maml{} and \norml{}.

\section{Experimental Setup}

In this section, we describe the comparisons and implementation details of our experiments.

\subsection{Comparisons\label{legend_def}}
We compare \norml{} to two existing approaches: vanilla \maml{}~\cite{DBLP:journals/corr/FinnAL17} and Domain Randomization~\cite{sadeghi2016cad2rl,DBLP:journals/corr/TobinFRSZA17,peng2018sim} (DR).
Domain randomization aims to learn a single robust policy by varying the environment for each rollout. We implement domain randomization by setting the adaptation learning rate $\alpha$ and the policy offset $\bm{\theta}_{\small\mbox{offset}}$ to zero (and disabling meta learning of these parameters) in our \maml{} implementation. Hence the meta-policy is directly used to compute the average loss across tasks/randomizations. This eliminates other factors that could influence experimental results and ensures that we are doing a fair comparison.

In addition, we also perform an ablation study for different components of \norml{}. We refer to them as \textit{\norml{} w/o offset}, which uses a learned advantage function but does not include offset as a trainable parameter, and \textit{\norml{} w/o LAF}, which, like MAML, uses ground-truth external reward but also includes the offset.

For a fair comparison, all algorithms are trained for the same number of iterations and with the same number of timesteps collected per iteration. 
For all experiments, we randomly sweep the following three hyperparameters: the outer learning rate $\beta$, the adaptation learning rate $\alpha$, and the initial value of the policy standard deviation $\bm{\theta}_{\sigma}$. We then plot the learning curves using the best hyperparameters found.  

\subsection{Implementation Details\label{section:implementation_details}}
We represent our policy as a multivariate diagonal Gaussian distribution and use a fully-connected, feed-forward network to map states $\bm{s}_t$ to a distribution over action $\pi_{\bm{\theta}}(\bm{a}_t|\bm{s}_t)$. 
The neural network outputs the mean of the Gaussian policy, and we used standalone variables to represent the standard deviations of each dimension: $\pi_{\bm{\theta}}(\bm{a}_t|\bm{s}_t)=\mathcal{N}(f(\bm{s}_t |  \bm{\theta}_{\mu}), \mbox{diag}(e^{\bm{\theta}_{\sigma}})^2)$ ($\bm{\theta}=\left[\bm{\theta}_{\mu}^T\,\,\bm{\theta}_{\sigma}^T\right]^T$). We found this to greatly improve training stability, compared to having the network output both the mean and log standard deviation.
We use a two-layer fully connected network with $\tanh$ activation function for the policy network ($50$ neurons per layer).
Similarly, the learned advantage function $\adv$ uses a fully-connected, two-layer neural network with rectifying linear units ($50$ neurons per layer).

For our \maml{} implementation, we also included the Meta-SGD extension by \citet{DBLP:journals/corr/LiZCL17MetaSGD}, which replaces the fixed inner learning rate of \maml{} with a learned vector of the same dimension as the policy parameter. 
In practice, hyperparameter optimization of the initial value of $\bm{\alpha}$ is required as tuning learning rates is a much slower process than optimizing the meta-policy $\bm{\theta}$. We used this extension in our experiments to improve the expressiveness of \maml{}'s adaptation step.

For meta training, we use Proximal Policy Optimization~\cite{DBLP:journals/corr/SchulmanPPO17} (PPO) for \maml{}'s adaptation and meta objectives to improve performance. For \norml{} PPO is only used for the meta objective as the learned advantage function $\adv$ learns to transform the vanilla policy gradient. 
As in the original \maml{} paper~\citep{DBLP:journals/corr/FinnAL17}, we use polynomial regression per task to fit the value function~\cite{duan2016benchmarking} (in both the adaptation and meta learning steps for \maml{} and only in the meta learning step for \norml{}). We use Adam~\cite{DBLP:journals/corr/KingmaB14} as our meta-policy optimizer and use vanilla policy gradient in the inner loop to avoid over-complicating the meta objective.
To speed up computation, our \maml{} and \norml{} implementations are parallelized across tasks and rollouts.

For all of our experiments, we use $K=25$ rollouts for adaptation and meta learning. We sample $10$ tasks $\mathcal{T}_i$ during each meta-training iteration.

\section{Point Agent Case Study}
\begin{figure*}[htp]
        \centering
        \begin{subfigure}[b]{0.4\textwidth}
            \centering
            \includegraphics[width=\textwidth]{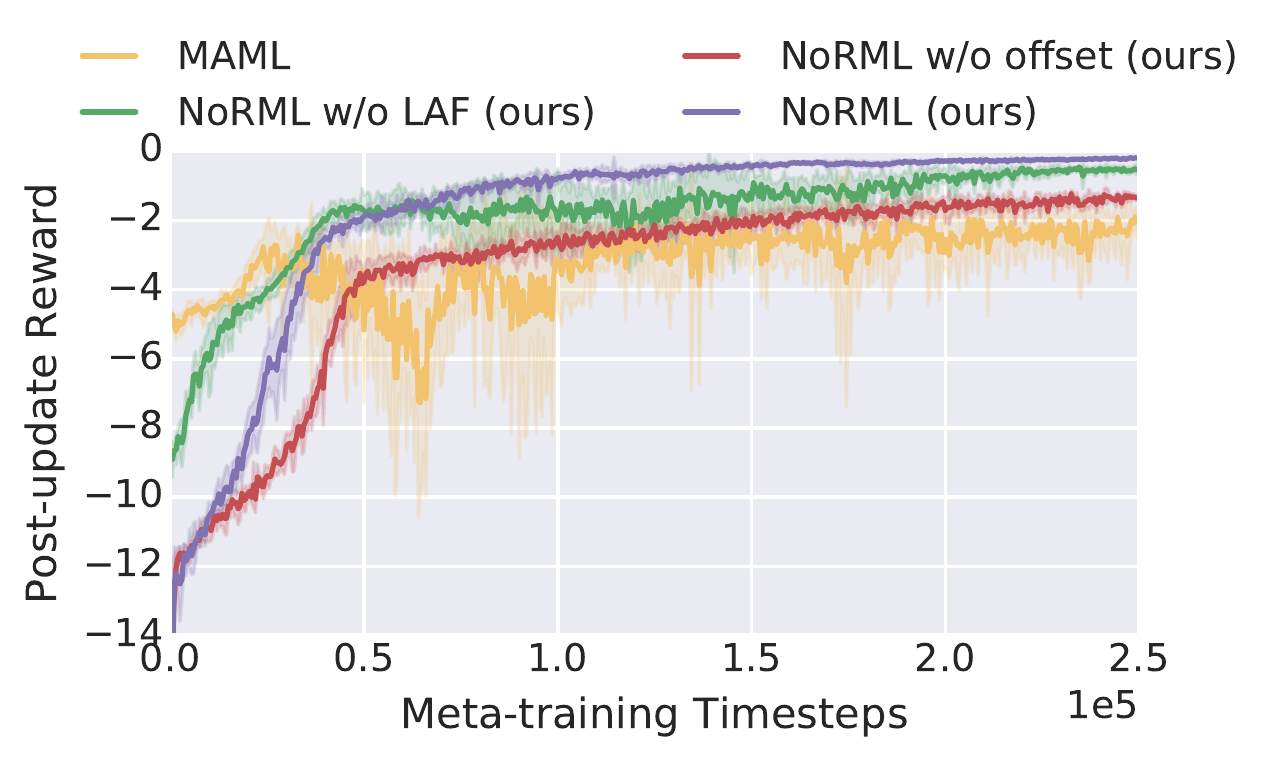}
            \caption
            {{\small Shaped Reward: Learning Curve}}    
            \label{fig:pointmass_continuous_learning_curve}
        \end{subfigure}
        \begin{subfigure}[b]{0.4\textwidth}  
            \centering 
            \includegraphics[width=\textwidth]{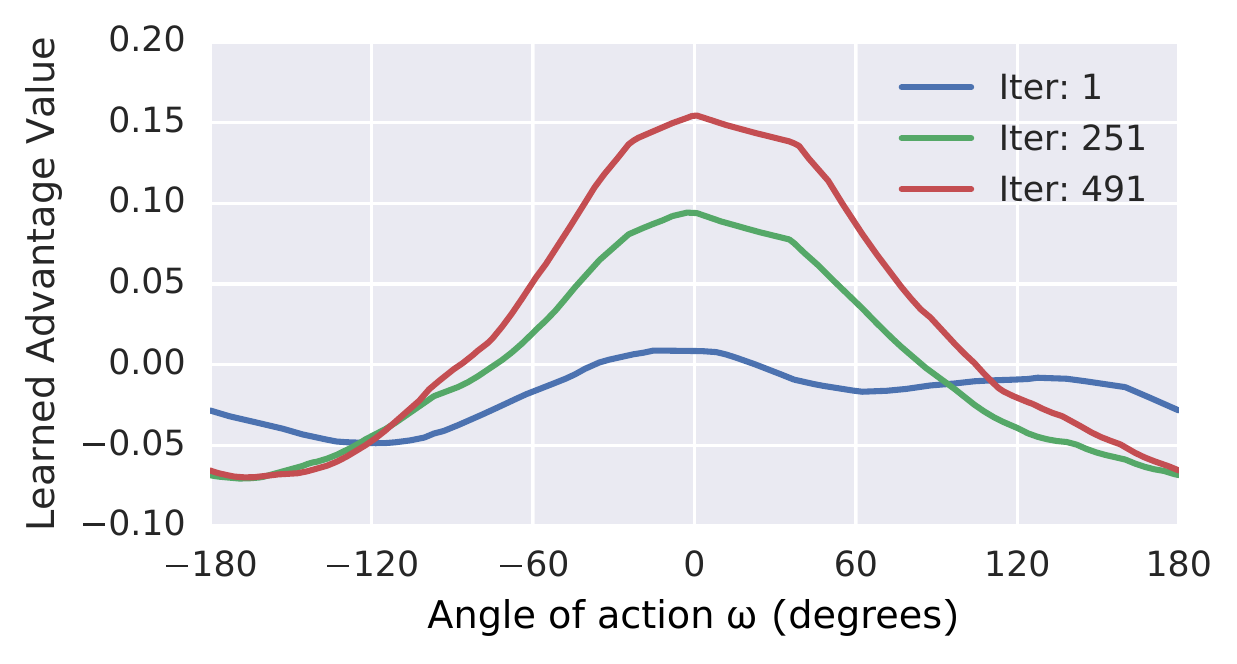}
            \caption[]%
            {{\small Shaped Reward: Learned Advantage Function}}    
            \label{fig:pointmass_continuous_advantage_function}
        \end{subfigure}
        \begin{subfigure}[b]{0.4\textwidth}   
            \centering 
            \includegraphics[width=\textwidth]{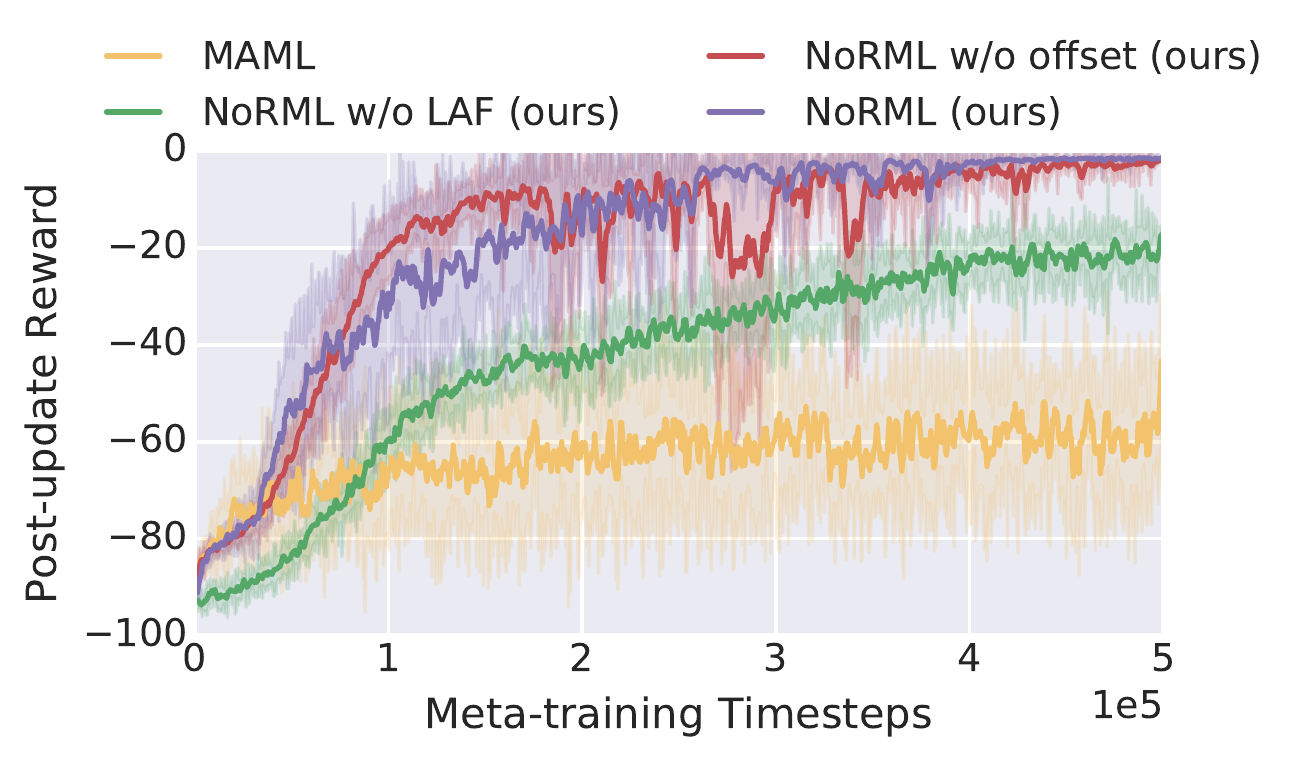}
            \caption[]%
            {{\small Sparse Reward: Learning Curve}}    
            \label{fig:pointmass_discrete_learning_curve}
        \end{subfigure}
        \begin{subfigure}[b]{0.4\textwidth}   
            \centering 
            \includegraphics[width=\textwidth]{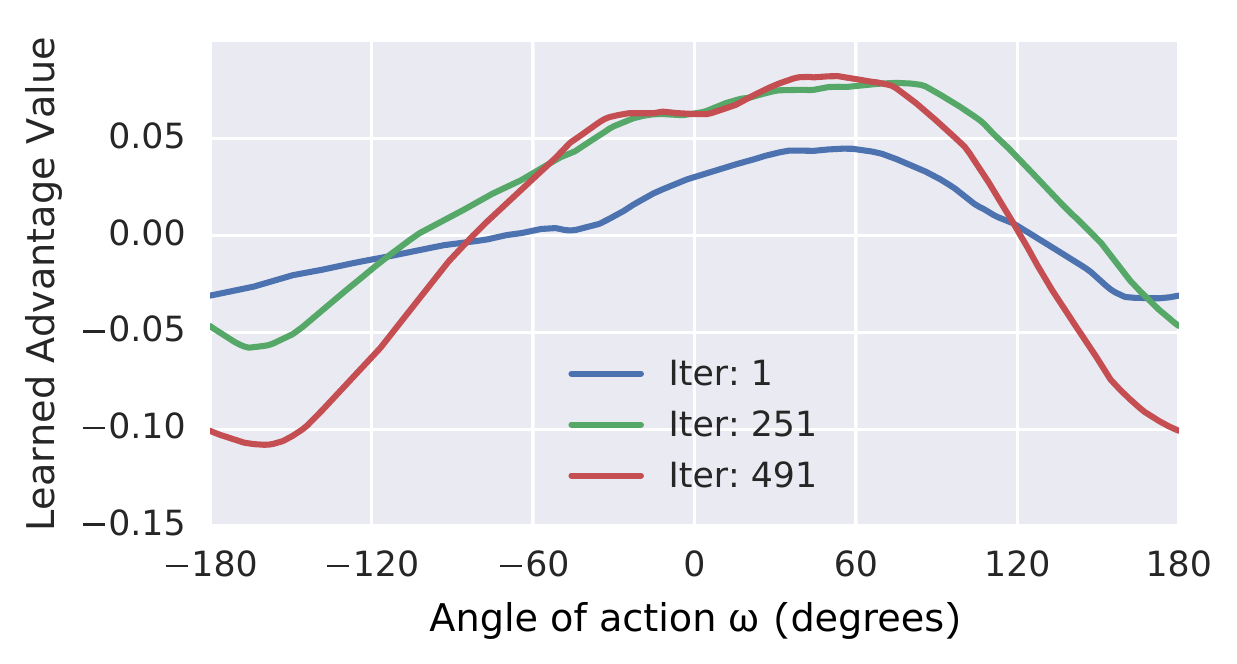}
            \caption[]%
            {{\small Sparse Reward: Learned Advantage Function}}    
            \label{fig:pointmass_discrete_advantage_function}
        \end{subfigure}
        \caption
        {Point agent: Meta-training curves and the learned advantage function for held-out tasks. LAF means "learned advantage function" (see section~\ref{legend_def} for details). The advantage values are plotted on actions of length 1 on origin $(0, 0)$, evaluated on the task with no rotation ($\phi=0$). Therefore, an action at angle $0\degree$ takes the point to the goal $(1, 0)$, and an action at angle $180\degree$ takes the point to $(-1, 0)$. NoRML is able to learn a shaped advantage function that leads to effective adaptation to dynamics, even when only sparse rewards are provided.}
        \label{fig:mean and std of nets}
\end{figure*}
\begin{figure}[htp]
    \centering
    \includegraphics[width=0.4\textwidth]{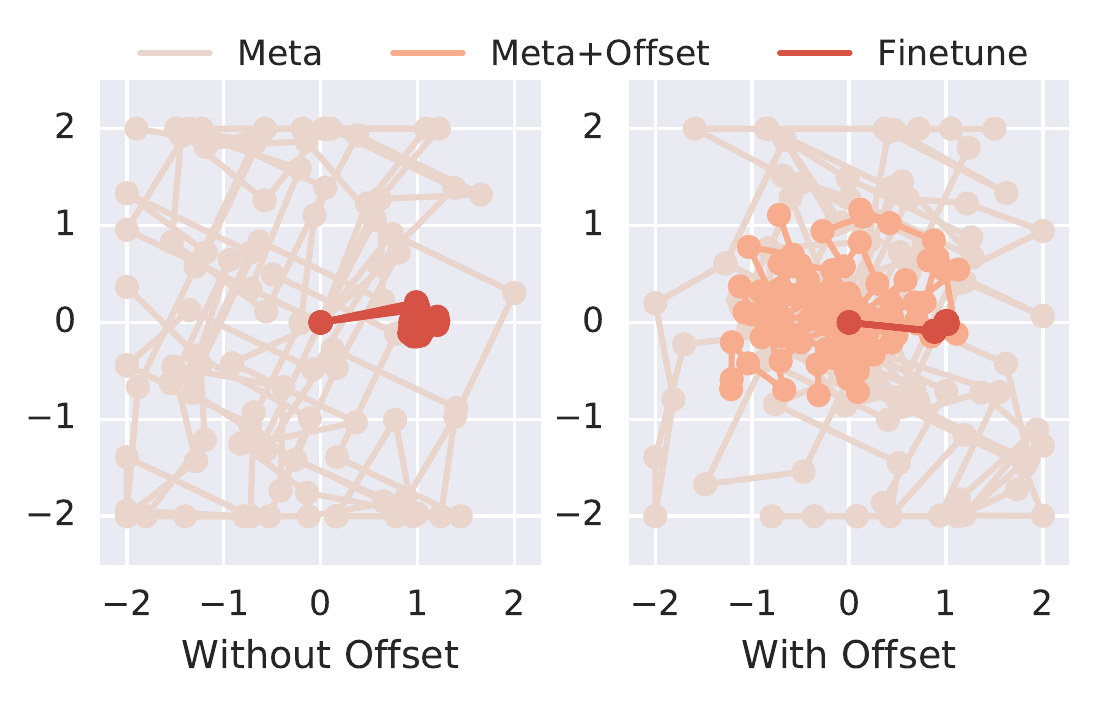}
    \caption{\label{fig:pointmass_offset}10 rollout trajectories for \norml{} policies trained with and without the offset. "Meta+Offset" means we only add the learned offset $\bm{\theta}_{\tiny\mbox{offset}}$ to the meta-policy parameters $\bm{\theta}$ without a gradient step and evaluate policy $\pi_{\bm{\theta}+\bm{\theta}_{\tiny\mbox{offset}}}$.}
\end{figure}
We now introduce a simple, 2D point agent control problem to illustrate the advantage function learned by \norml{} and study its effect with both shaped and sparse rewards. For this simple, synthetic control task, we show that \norml{} learns an intuitive, realistic advantage function and achieves similar performance to \maml{}. As we increase the task's difficulty by making the reward sparse, \maml{}  struggles to learn, while \norml{} still learns a similar advantage function that allows it to adapt to dynamics changes.
\subsection{Task Setup}
Consider an agent in a plane that is trying to move to the right from $(0, 0)$ to $(1, 0)$. The agent observes its current position $\observation_t=\left[x_t\, y_t\right]^T$, and each action specifies its movement $\action_t=\left[dx_t, dy_t\right]^T$, where $dx_t, dy_t\in\left[ -1, 1\right] $. We introduce a dynamic change where the action is rotated by $\phi$ (one $\phi$ per task) that is unknown to the agent. The new dynamics can be expressed as:
\begin{equation}
    \left[\begin{array}{c}
         x_{t+1}\\y_{t+1}
    \end{array}\right]=
    \left[\begin{array}{cc}
       \cos\phi  & -\sin\phi \\
       \sin\phi  & \cos\phi
    \end{array}\right]\left[\begin{array}{c}
         dx_t\\dy_t
    \end{array}\right]+\left[\begin{array}{c}
         x_t\\y_t
    \end{array}\right].
\end{equation}
We also restrict the agent's movement to a square region {$-2 \leq x_{t+1}, y_{t+1} \leq 2$}.

We consider two types of reward functions: shaped and sparse. In the former case, each rollout has a fixed horizon of $10$ steps, and the reward at each step is the negative Euclidean distance to the destination $(1, 0)$. For the sparse reward case, the reward is $-1$ for each step taken, but an episode can terminate early when the agent successfully reaches within 0.1 radius to the goal. Each rollout has a maximum horizon of $100$ steps.

As a meta learning problem, different tasks are defined by the unknown rotation $\phi$, where the task distribution is uniform on $[0,2\pi]$ radians. The agent needs to gather information about the rotation amount $\phi$ in the meta rollouts, and make corresponding changes to the policy during fine-tuning.

\subsection{Impact of the Learned Advantage Function}
To visualize the advantage function learned by \norml, we transformed the original action space to polar coordinates and plotted the learned advantage functions, as seen in fig.~\ref{fig:pointmass_continuous_advantage_function} and \ref{fig:pointmass_discrete_advantage_function}. In both cases, the agent is located at the origin and takes an action of length 1 and angle $\omega$, which would move the agent from $\left[0, 0\right]^T$ to $\left[\cos\omega, \sin\omega\right]^T$. Therefore, an action with angle $\omega=0$ would take the agent directly to the desired destination.

In the shaped-reward case (Fig.~\ref{fig:pointmass_continuous_advantage_function}), we see that the learned advantage function gradually converges to a bell-like shape peaked around $0\degree$, which rewards actions that move the agent towards the destination. Note that the advantage network only takes in $\transition$ as  input and does not have access to the ground-truth reward value. After meta training, it is able to learn a smooth advantage function that guides the meta-policy for proper fine-tuning.

As we make the task harder by introducing the sparse reward function, \norml{} still learns a similarly-shaped reward function shown in Fig.~\ref{fig:pointmass_discrete_advantage_function} with peak value around angle $0\degree$. In this more difficult task, \maml{}'s performance degrades dramatically (Fig.~\ref{fig:pointmass_discrete_learning_curve}). The sparse reward function is challenging for vanilla \maml{} to adapt to changes in dynamics, since it uses a single policy gradient step with a fitted value function. In contrast, the learned advantage function in \norml{} provides a more informative reward signal that enables the agent to adapt in one policy gradient step.

\subsection{Impact of the Offset}
We find that the parameter offset $\bm{\theta}_{\tiny\mbox{offset}}$ encourages exploration in \norml{}. In Fig.~\ref{fig:pointmass_offset}, we train two \norml{} policies, one with and one without a policy offset and we plot the trajectories sampled from these policies. The meta-policies in both cases tend to be exploratory and have larger variance, and the fine-tuned policies are more consistent and move directly to the destination. However, with offset enabled, the fine-tuning process in Fig.~\ref{fig:pointmass_offset} learns to reduce variance even further, since the offset already reduces the explorative policy to a more conservative one.

The effect of the offset is further illustrated in Fig.~\ref{fig:pointmass_continuous_learning_curve}, where the offset allows the fine-tuned policy to converge to better final values. Without the offset, both \maml{} and \norml{} could not achieve a total reward greater than $-1$. When the policy offset is enabled, however, the learning curves converged to values as high as $-0.4$, which is a significant improvement for this task.

\section{Continuous Control Tasks}
To study how \norml{}  scales to more complex deep RL problems, we apply it to two continuous control problems in the OpenAI Gym~\cite{openaigym}, and compare it with vanilla MAML and domain randomization.

\subsection{Cartpole with Sensor Bias}
\begin{figure}[htp]
        \centering
        \begin{subfigure}[b]{0.4\textwidth}
            \centering
            \includegraphics[width=\textwidth]{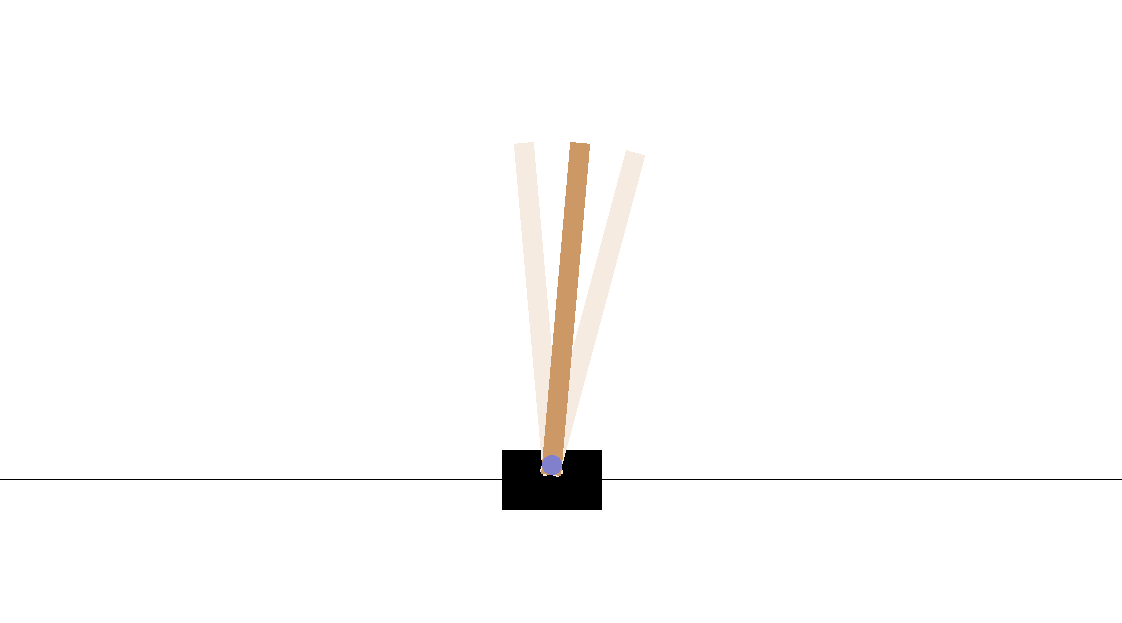}
            \caption
            {{\small The cartpole's uncalibrated angle sensor adds an unknown bias to the actual reading. When the pole is at the  position shown in the figure, the sensor's reading can indicate any reading between the two blurred ones.}}    
            \label{fig:cartpole_concept}
        \end{subfigure}
        \begin{subfigure}[b]{0.4\textwidth}  
            \centering 
            \includegraphics[width=\textwidth]{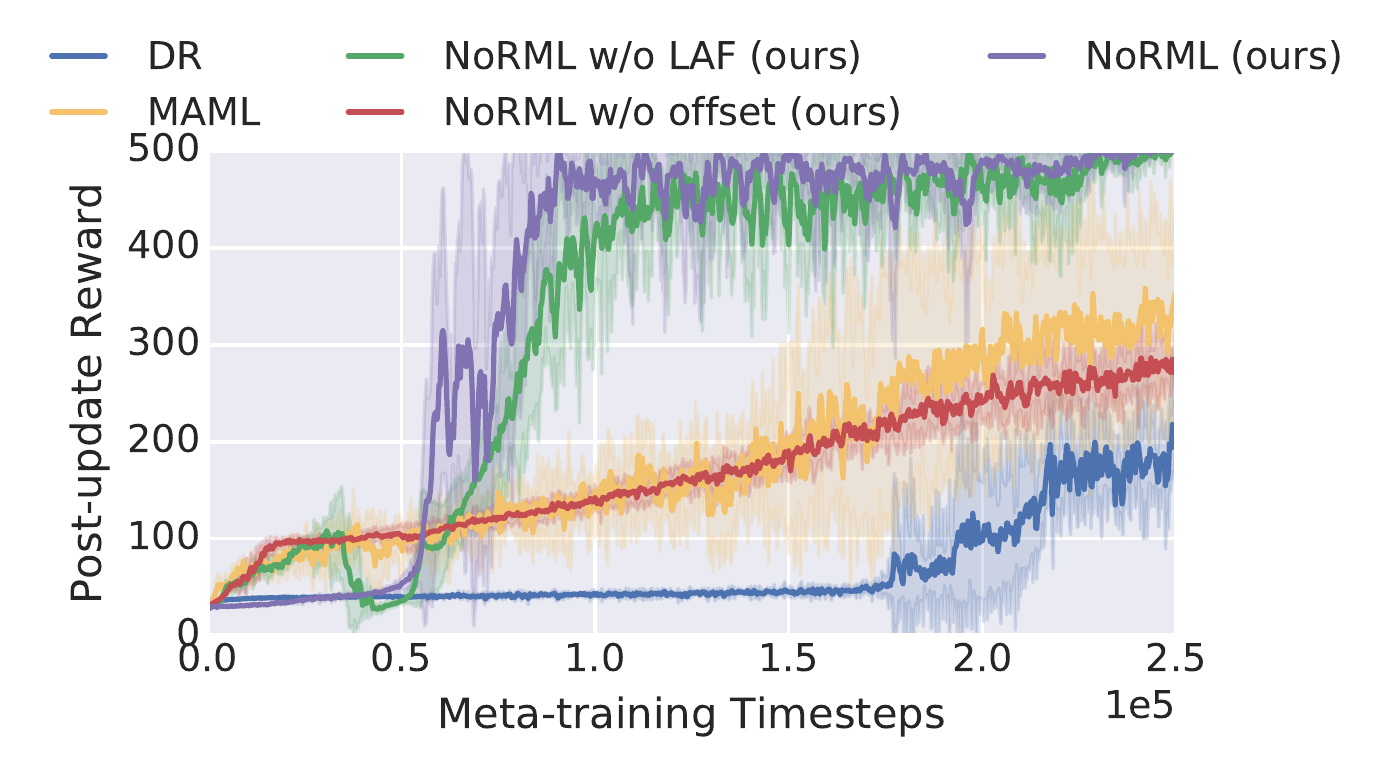}
            \caption[]%
            {{\small Meta-training curve on held-out tasks: since the reward is $1$ for every time step that the pole does not fall down, the total reward reflects the length of the episode and the maximum reward is $500$. Note that \norml{} only uses the cart and pole's position and velocity during adaptation, while \maml{} additionally uses external rewards.}}    
            \label{fig:cartpole_learning_curve}
        \end{subfigure}
        \caption
        {Illustration of the cartpole task and results. We find that adaptation is critical for this task, that our method converges faster, and that both the learned advantage and learned offset are helpful.}
        \label{fig:cartpole}
\end{figure}

We introduce a variant of the Cartpole environment, in which the agent needs to move the cart to balance an inverted pendulum. For each time step, the agent observes the position and velocity of both the cart and the pole, and applies a force to the cart in order to balance the system. In our variant, the position sensor can drift: its reading can be offset by an unknown amount ranging from $-10\degree$ to $10\degree$ (Fig.~ \ref{fig:cartpole_concept}). Hence, our meta-training task distribution corresponds to a uniform distribution over this range of sensor reading. We also make the task more difficult by increasing the required duration of balancing the pole from $4$ seconds, as in the original OpenAI Gym environment, to $10$ seconds.

As seen in Fig.~\ref{fig:cartpole_learning_curve}, although both vanilla \maml{} and \norml{} converge to a reward of 500 in the end, \norml{} converges faster despite having fewer assumptions---\norml{} does not require an external reward signal for its adaptation.
Domain randomization cannot solve the task in this case, demonstrating that adaptation is necessary to solve these tasks. The plot also shows an ablation study of different components: without the offset, \norml{} could not converge to a high final reward, and without the learned advantage function, convergence is slower.

\subsection{Half Cheetah with Swapped Actions}
\begin{figure}[htp]
    \centering
    \begin{subfigure}[b]{0.45\textwidth}
        \includegraphics[width=\textwidth]{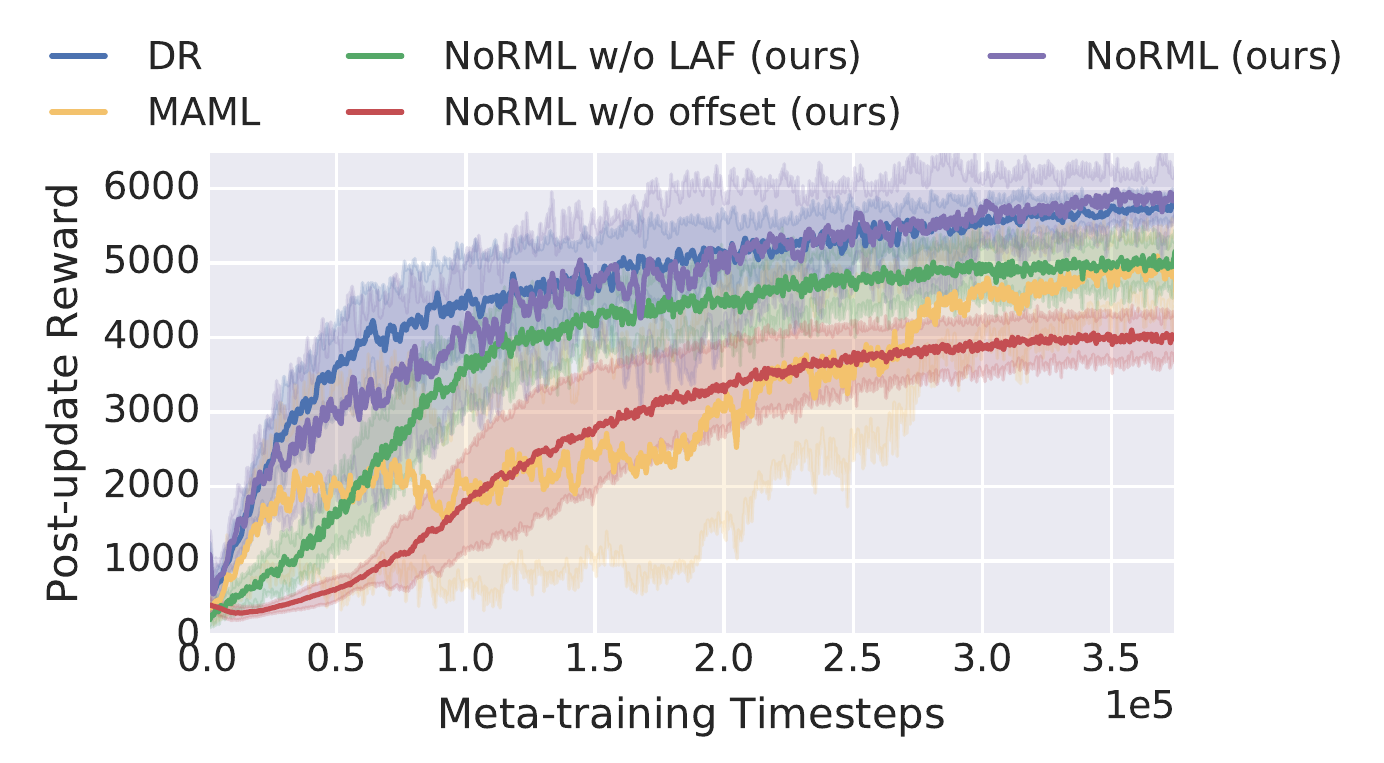}
        \caption{\small Meta-training curve for the half cheetah. In this case, MAML struggles to recognize and adapt to dynamics changes, while the learned advantage function enables effective adaptation. Domain randomization also leads to high reward, but cannot produce as stable of a gait (see below).}
        \label{fig:cheetah_learning_curve}
    \end{subfigure}
    \begin{subfigure}[b]{0.4\textwidth}
        \centering
        \includegraphics[width=0.55\textwidth]{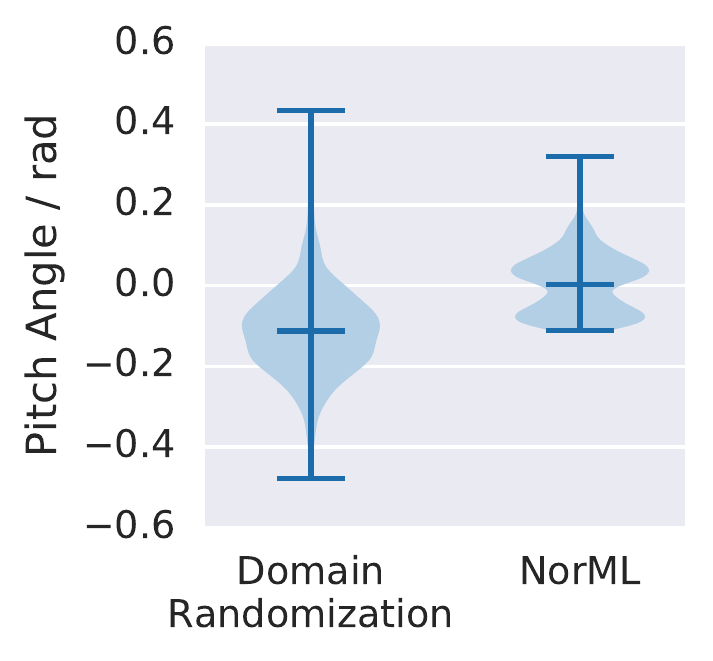}
        \caption{\small Pitch  distribution showing increased stability of the fine-tuned \norml{} gait compared to the gait learned with Domain Randomization.}
        \label{fig:cheetah_imu}
    \end{subfigure}
    \begin{subfigure}[b]{0.45\textwidth}
        \includegraphics[width=\textwidth]{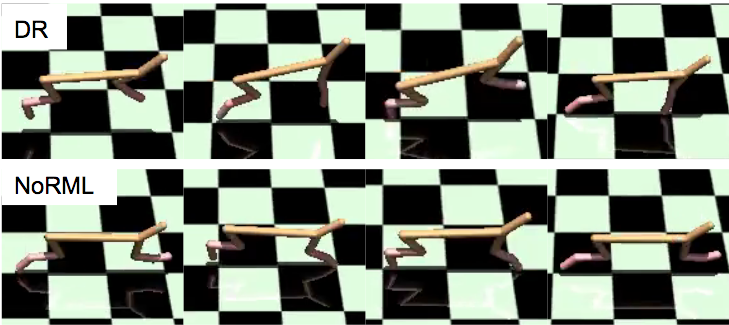}
        \caption{\small Gait learned by Domain Randomization (top) and \norml{} (bottom). The fine-tuned gait learned by \norml{} is more stable and the body oscillates less.}
        \label{fig:cheetah_gait}
    \end{subfigure}
    \caption{Learning curve, IMU readings and snapshots of the running gait for the HalfCheetah environment.}
\end{figure}

We next evaluate on the half cheetah environment in OpenAI Gym. To test \norml{}'s adaptability to dynamic changes, we purposefully allow the torque outputs of the two hip joints to be swapped, leading to two different tasks. In real robotic problems, this change could occur due to wiring and signal transmission errors. 
Another challenge of robot locomotion in the real world is the unavailability of accurate, real-time on-board localization systems. To simulate this, we remove the position and linear velocity from half cheetah's observation space. Note that the lack of localization information also makes it difficult to compute the distance-based reward function. Without reward function, \maml{} can no longer perform adaptation, while \norml{} can still adapt using the learned advantage.

With the learned advantage function, \norml{} significantly outperforms vanilla MAML both in convergence speed and final return (Fig.~\ref{fig:cheetah_learning_curve}). Moreover, although domain randomization achieved similar performance in terms of final return, we found the gait learned by domain randomization to be less stable (Fig.~\ref{fig:cheetah_imu}): the body of the half cheetah oscillates a lot during running (Fig.~\ref{fig:cheetah_gait}), as domain randomization must learn a single policy that can handle all action swaps.
On the other hand, with \norml{}, the policy can adapt quickly and gracefully, without reward or tracking information, regardless of the actions being swapped.

\begin{figure}[htp]
    \centering
    \includegraphics[width=0.45\textwidth]{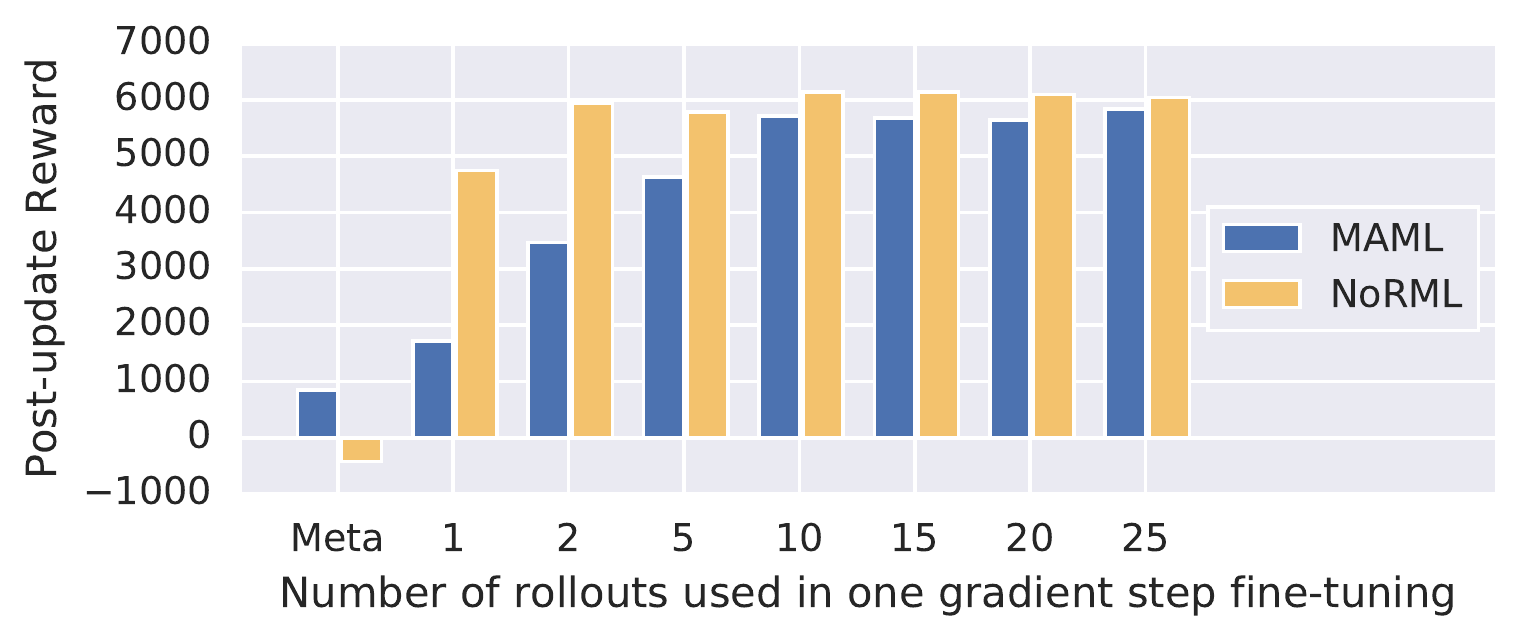}
    \caption{Post-update rewards for HalfCheetah using variable numbers of rollouts. \norml{} can adapt effectively with as few as one roll-out, while \maml{} cannot adapt well using fewer than 5 rollouts.}
    \label{fig:cheetah_rollouts}
\end{figure}

For the HalfCheetah policy, we also took the top-performing policy and tested its adaptation performance using a smaller number of rollouts. As shown in Fig.~\ref{fig:cheetah_rollouts}, even with as little as a single rollout, the fine-tuned policy still achieves a reasonable performance. 
This shows that the learned advantage function is noise-tolerant and can sense dynamics changes using a small amount of sample data, as little as a single trajectory. On the other hand, \maml{} needs at least 5 meta rollouts to achieve a similar post-adaptation performance, due to the noisy estimation of value function and advantage.

\section{Conclusion}
In this paper, we introduce \norml{}, a new meta reinforcement learning algorithm that adapts to changes in dynamics and sensor drifts, without the need for external reward signals during adaptation. 
The key insight is to learn an advantage function, a parameter offset between meta and adapted policies, and the meta policy simultaneously. 
The learned advantage function results in a more expressive adaptation step by generalizing \maml{}'s update rule. 
The parameter offset encourages exploration during meta rollouts by decoupling the fine-tuned policies from the meta-policy. 

We evaluate \norml{} on three control problems with varying types of dynamics changes: an illustrative point agent example with distorted actions, a cartpole with sensor bias, and a half-cheetah with wiring errors. 
Our experiments show that, by incorporating both a learned advantage and a learned offset, \norml{} can adapt to all these types of changes, even when the reward signals are not present during adaptation. 

A promising future research direction is to apply \norml{} to transfer policies from simulation to real robots. Sim-to-real transfer is a challenging problem in robotics, which is caused by model errors between the simulation and the real-world physics. We can treat the model error as the dynamics change and apply \norml{} to adapt to this change with a few shots of real robot data.

\balance
\bibliographystyle{ACM-Reference-Format}  %
\bibliography{sample-bibliography}  %

\end{document}